\documentclass[12pt,a4paper]{cibb}

\usepackage{subfigure,graphicx}
\usepackage{amsmath,amsfonts,latexsym,amssymb,euscript,xr}
\usepackage{booktabs}
\usepackage[nodayofweek]{datetime}
\usepackage{hyperref}

\usepackage[table]{xcolor}
\usepackage{color,colortbl,tabularx}

\usepackage[english]{babel}
\usepackage[protrusion=true,expansion=true]{microtype}
\usepackage{amsmath,amsfonts,amsthm}
\usepackage{pifont}
\usepackage{multirow}
\usepackage{makecell}

\definecolor{LightBlue}{rgb}{0.88,0.9,0.9}

\title{\Large $\ $\\ \bf MIEO: encoding clinical data to enhance cardiovascular event prediction}

\author{\large Davide Borghini$^1$, Davide Marchi$^{1}$, Angelo Nardone$^{1}$, Giordano Scerra$^{1}$, Silvia Giulia Galfrè$^{1}$, Alessandro Pingitore$^2$, Giuseppe Prencipe$^{1}$, Corrado Priami$^{1}$, and Alina S\^irbu$^{*,3}$}
\address{\footnotesize $\ $\\$^1$ Department of Computer Science, University of Pisa,
Pisa, Italy. d.borghini3@studenti.unipi.it, d.marchi5@studenti.unipi.it, a.nardone5@studenti.unipi.it, g.scerra1@studenti.unipi.it, silvia.galfre@di.unipi.it, giuseppe.prencipe@unipi.it,  corrado.priami@unipi.it  \\
$^2$ Clinical Physiology Institute, CNR, Pisa, Italy. alessandro.pingitore@cnr.it \\ %
$^3$ Department of Computer Science and Engineering, University of Bologna,
Bologna, Italy. alina.sirbu@unibo.it %
\bigskip
$^*$corresponding author
}

\abstract{\small clinical data, cardiovascular disease, autoencoder, neural network, event prediction. \normalsize
\\[17pt]
{\bf Abstract.} As clinical data are becoming increasingly available, machine learning methods have been employed to extract knowledge from them and predict clinical events. While promising, approaches suffer from at least two main issues: low availability of labelled data and data heterogeneity leading to missing values. This work proposes the use of self-supervised auto-encoders to efficiently address these challenges. We apply our methodology to a clinical dataset from patients with ischaemic heart disease. Patient data is embedded in a latent space, built using unlabelled data, which is then used to train a neural network classifier to predict cardiovascular death. Results show improved balanced accuracy compared to applying the classifier directly to the raw data, demonstrating that this solution is promising, especially in conditions where availability of unlabelled data could increase. 
}

\begin{document}
\thispagestyle{myheadings}
\pagestyle{myheadings}
\markright{\tt Proceedings of CIBB 2025}

\section{\bf Introduction}
\label{sec:SCIENTIFIC-BACKGROUND}

The application of machine learning (ML) to clinical data has gained increasing attention in recent years, particularly in the context of predicting cardiovascular events \cite{goldstein2017opportunities, choi2017using, pingitore2024machine}. Electronic health records contain large amounts of patient information that can be leveraged to support clinical decision-making. However, the successful application of ML to such data is limited by several challenges, among which two are especially prominent: (i) the scarcity of labelled clinical datasets, and (ii) the heterogeneity of the data, often leading to a high proportion of missing values.

In this work, we address these challenges through the introduction of MIEO (Masked Input Encoded Output), a self-supervised autoencoder model, designed for \emph{structured clinical data}. The model takes advantage of unlabelled clinical data, which is typically more easily available compared to labelled data, while handling missing values explicitly. We train our model on data from patients with ischaemic heart disease, and test the embeddings obtained on a downstream task: prediction of cardiovascular death. The MIEO approach obtains better balanced accuracy compared to classifying patients using  their clinical data features directly.

\section{\bf Data and Methods}
\label{sec:DATA-AND-METHODS}

\subsection{Data}
The study included 8065 IHD patients hospitalized (1977–2011) at the CNR Clinical Physiology Institute in Pisa, Italy. The dataset comprises 68 variables, each representing a specific clinical attribute. Among these, 46 columns are binary variables, indicating conditions such as ``Smoke'' ``Diabetes'' and ``Myocarditis''. The remaining columns represent continuous variables, including ``Creatinine'', ``Vessels'' and ``Total Cholesterol''. It represents an extension of the dataset used in~\cite{pingitore2024machine}.

The data went through a preprocessing phase where missing values and outliers were set to null. After these steps, the dataset had a total of 2.98\% missing data. In particular, most features had no missing values, a small portion had approximately 5\% missing data, and only two features had more than 50\% missing data. All features were maintained in the dataset, including the two with more than 50\% missing values. 

The classification target of our study was cardiovascular death (CVD) within eight years from the first visit. We therefore labelled our patients with two labels: 0 if they survived more than 8 years and 1 if they had a CVD event within that time.  Some patients could not be labelled: if they had the final follow-up or a death event that was not cardiovascular before 8 years. We therefore obtained two datasets: a \emph{labelled} dataset of 3769 patients and an \emph{unlabelled} one of 4296 patients. 

Our data underlines the challenges in applying ML techniques to clinical data: features with missing data are always present, and a large amount of patients may have missing followup resulting in unlabelled data that cannot be used for classification.

The labelled data was split into three subsets: 20\% of patients belonged to the test dataset, and 80\% to the development dataset. The development dataset was further split into train (80\%) and validation (20\%).   

\subsection{Model}
\begin{figure}[h!]
    \centering
    \includegraphics[scale=0.5]{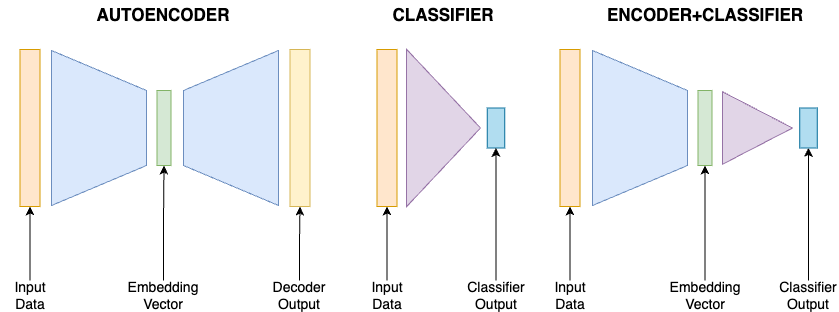}
    \caption{A simplified representations of our models: the MIEO autoencoder, the classifier for the downstream task applied to clinical data directly, the classifier applied to MIEO embeddings.}
    \label{fig:models}
\end{figure}

The main model we introduce is the MIEO model (Masked Input Encoded Output), a modified autoencoder that maps into a latent space unlabelled clinical data with missing values. To evaluate the usefulness of the embeddings obtained, we apply them to a downstream classification task: predicting CVD within 8 years, using an artificial neural network model (ANN). We compare the prediction performance of the ANN model with and without the use of embeddings (applied directly on the clinical variables). Figure \ref{fig:models} provides a simplified representation of the methodology used.  In the rest of this section we detail the architecture of the various models.

\subsubsection{MIEO autoencoder}

MIEO  was designed by modifying the classic autoencoder structure to handle null values and also perform imputation on these data, while optimizing the handling of binary data. To achieve this, we used two main components: a mask integrated into the input and a custom loss function adapted to our problem. 

\begin{figure}[h]
    \centering

    {\includegraphics[scale=0.35]{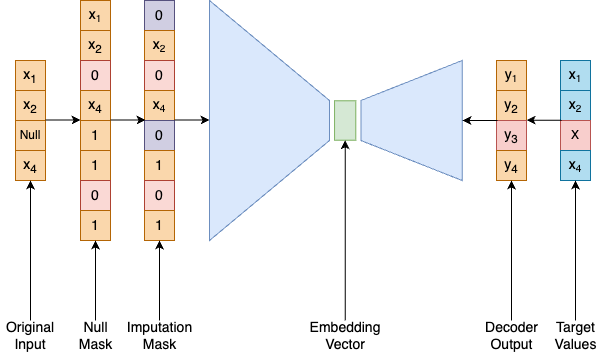}} \quad
    \caption{ A graphical example of the target value and the output used to calculate the loss in the MIEO model.}
    \label{fig:MIEO model}
\end{figure}

The mask indicates which features in the input have missing values. We add further missing values to the data (i.e. masking some additional features), thus also generating additional data for training. The autoencoder is fed the masked input data and asked to reproduce the unmasked one. This operation helps with missing data imputation. Since the unmasked data could still contain  null  (missing) values, we compute the loss function only on the non-null ones.  Figure \ref{fig:MIEO model} provides a visual representation of how our model operates to generate the output.

In computing the loss, our approach treats entries with binary versus continuous data differently. For binary data we use the Binary Cross-Entropy (BCE), while for continuous data we use the Mean Squared Error (MSE). These are combined linearly, using two weights (which become model hyperparameters), due to the prevalence of the binary features. Giving greater weight to the continuous loss will make the model focus more on improving the weights that lead to continuous data outputs, and vice versa. 

The model architecture is a  classical autoencoder. Specifically, we used four encoding layers to generate the embedding vector, followed by four decoding layers, producing the output vector. The size of the embedding vector, which is the dimension of the latent space for encoding, is a parameter of the model. We explore both small and large values, even larger than the input data, resulting thus in an expansion of the feature space.  We selected LeakyReLU as the activation function, as it provided the best results. To further enhance performance, we applied Batch Normalization at the end of each layer, to normalize output and improve training stability~\cite{ioffe2015batch}.

 In summary, these two modifications (input masking, and loss function differentiated between discrete and continuous variables) should enhance our model compared to a standard autoencoder, making it capable of imputing null data when working with new, unseen data and optimally handling the presence of a large amount of binary data.

\subsubsection{ANN Classifier Model}\label{classmodel}

For the downstream classification task, i.e. prediction of the CVD label, we employed a standard feedforward neural network classifier, with three LeakyReLU hidden layers and Batch Normalisation. This model was applied in two scenarios: in the first, it directly processed the original input data, while in the second, it received input data projected into a latent space using the encoder trained with the MIEO model presented in the previous section. 

We used the standard Binary Cross-Entropy loss with positive weights, which simulates an oversampling process. We chose this approach due to the class imbalance in the target distribution, aiming to balance the future results across both classes. The only differences between the two classifiers are as follows:
\begin{itemize}
    \item ANN Only: In this case, we applied the Null Mask to the original data, ensuring that missing data did not affect the classification task.
    \item MIEO+ANN: We used input data from the latent space generated by the MIEO model. Therefore, no mask was required for these data. 
\end{itemize}

Since the input data differed in these two cases, we performed separate hyperparameter optimisation through grid search.

\section{\bf Results}
\label{sec:RESULTS}

The MIEO autoencoder was trained using both \emph{labelled} data from the \emph{training} dataset and the \emph{unlabelled} data. Both ANN models were trained using labelled training data, ensuring a fair comparison between the approaches. The validation dataset was used to optimise ANN and MIEO hyperparameters. The test dataset was finally employed to test the final model obtained.

Selecting the MIEO model solely based on its ability to reconstruct the input with the highest accuracy did not prove to be the best approach. This was primarily due to two factors: the presence of missing data and the imbalance between binary and continuous features. Since our dataset contained significantly more binary features than continuous ones, a standard autoencoder would tend to prioritize binary features during reconstruction.  To address this issue, we generated and trained a large number of MIEO models, each with different weights and hyperparameters, using a standard grid search. However, instead of selecting the best model based on reconstruction accuracy at this stage, we deferred the decision, and selected the encoding that produced best validation results in the downstream task. 

We used balanced accuracy both during hyperparameter optimisation and for model testing. This was due to the fact that, with our class imbalance problem, models tend to have low recall on the positive class (i.e. in recognising CVD events). We however aim to maximise this ability. 

Table~\ref{tab:results} shows the performance on validation and test data of the best model. We note similar performance when moving from validation to test, indicating promising generalisation abilities. Importantly, when looking at performance measures that take into account class imbalance, the model using the MIEO embeddings seems to perform slightly better: while macro average F1-scores are similar between the two ANN models, balanced accuracy (macro average recall) are better in the MIEO+ANN model. This indicates that the model is better able to recognise CVD events, although at the cost of a lower recall on class 0, and lower average precision. Overall accuracy is higher for the ANN model, however this is not relevant since accuracy is strongly influenced by class imbalance. Besides the slightly better recall of the downstream classifier, this result also indicates that the MIEO model is effective in extracting meaningful features from masked data, allowing the classifier that uses the latent space embeddings to match and slightly outperform the ANN trained directly on clinical data. Interestingly, the selected autoencoder employed an embedding size larger than the input layer.

\begin{table}[h!]
    \centering
    \begin{tabular}{|p{1cm}|c||c|p{1cm}|c||c|c|c||c|}
        \hline
        &  & \multicolumn{3}{c||}{MIEO +ANN} & \multicolumn{3}{c||}{ANN} &  \\

        \cline{2-9}
       & Class & Precision & \textbf{Recall} & F1 &Precision & \textbf{Recall} & F1 & Support \\
        \hline
        \multirowcell{5}{Valida-\\tion\\dataset}&0.0 & 0.908 & 0.790 & 0.845 & 0.867 & 0.898 & 0.882 & 472 \\
        &1.0 & 0.484 & 0.710 & 0.576 &0.579 & 0.5038 & 0.539 & 131 \\
        \cline{2-9}
        &Accuracy & & & 0.773 &  & & 0.813 &603 \\
        &\textbf{Macro avg} & 0.696 & \textbf{0.750} & 0.710 &0.723 & 0.701 & 0.711 & 603 \\
        &Weighted avg & 0.816 & 0.773 & 0.786 & 0.805 & 0.813 & 0.808 &603 \\
        \hline
        \hline
        \multirowcell{5}{Test\\dataset}&0.0 & 0.88 & 0.80 & 0.84 & 0.85 & 0.91 & 0.88 & 578 \\
        &1.0 & 0.49 & 0.65 & 0.56 &0.62 & 0.47 & 0.54& 176 \\
        \cline{2-9}
        &Accuracy & & & 0.76 & & & 0.81& 754 \\
        &\textbf{Macro Avg} & 0.69 & \textbf{0.72} & 0.70 &0.73 & 0.69 & 0.71 & 754 \\
       & Weighted Avg & 0.79 & 0.76 & 0.77 &0.80 & 0.81 & 0.80  & 754 \\
        \hline
       
    \end{tabular}
    \caption{Classification performance metrics on the  validation and test set for MIEO+ANN (applied on latent space features) and  ANN (applied directly to clinical features) classifiers. }
    \label{tab:results}
\end{table}

\section{\bf Conclusion}
\label{sec:CONCLUSIONS}

In this work, we proposed MIEO, a modified autoencoder architecture designed to handle missing values and heterogeneity in clinical data. Our model effectively integrates a masking mechanism and a differentiated loss function to process both binary and continuous features. These modifications allow for robust latent representations that are resilient to missing information and reflect clinically meaningful patterns.

We assessed the utility of the learned embeddings by using them in a downstream classification task for cardiovascular death prediction. Experimental results demonstrate that the MIEO+ANN approach achieves competitive, and in some metrics slightly superior, performance compared to a classifier trained directly on clinical variables. In particular, balanced accuracy improved with the use of latent space features, indicating a better sensitivity to CVD events in the presence of class imbalance.

The novelty of our contribution lies in the design of a self-supervised encoding method that can be trained on unlabelled data with missing values. This increases its applicability in real-world clinical settings, where labelled outcomes are scarce but patient data can be more abundant. Moreover, the mapping of patients in latent space can be seen as a novel way to design \emph{deep digital twins}, enabling the generation of synthetic yet realistic patient profiles for further simulation and predictive tasks. While the idea of a deep patient or deep digital twin is not new, the approach here is different from the majority of the literature that typically employs non-structured clinical notes to build concept embeddings~\cite{choi2016multi} or integrates structured and non structured clinical data to build patient profiles~\cite{miotto2016deep, gehrmann2018comparing, si2021deep}. We employ structured clinical data only, which can provide a more standardised way to integrate different datasets and could provide improved performance allowing to concentrate on specific variables, somehow similar to~\cite{wang2020application}. However, we specifically address missing data by including masking during self-supervised training. While our current model only produces a slight increase in performance compared to using clinical variables directly, we believe that by integrating large amounts of structured (unlabelled) clinical data we may be able to build advanced embeddings, which could ideally span different comorbidities. 

Future work will investigate the use of different models for the downstream task and the integration of additional data in the self-supervised phase of the analysis, which we expect should increase prediction power. A thorough analysis of imputation performance will also be performed. We will apply our approach to other clinical endpoints. Furthermore, different self-supervised architectures will be tested, and different ways of encoding input data with missing values.

\section*{\bf Conflict of interests}
\label{sec:CONFLICT-OF-INTERESTS}
The authors declare no conflicts of interest.


\section*{\bf Funding }
\label{sec:FUNDING}
This work was supported by the European Union - Next Generation EU, in the context of The National Recovery and Resilience Plan, Mission 4 Component 2, Investment 1.1, Call PRIN 2022 D.D. 104 02-02-2022 – MEDICA Project, CUP I53D23003720006; by the European Union – Horizon 2020 Program under the scheme “INFRAIA-01-2018-2019 – Integrating Activities for Advanced Communities”, project ``SoBigData++: European Integrated Infrastructure for Social Mining and Big Data Analytics'', Grant Agreement n.~871042; by the project Tuscany Health Ecosystem (THE), ECS00000017, Spoke 3, CUP: B83C22003920001 (European Union - Next Generation EU, in the context of The National Recovery and Resilience Plan, Investment 1.5 Ecosystems of Innovation);  by the University of Pisa through the SPARK project.  

\section*{\bf Availability of data and software code}
\label{sec:AVAILABILITY}
Our code is available on GitHub at the following link: \textcolor{blue}{\href{https://github.com/davide-marchi/clinical-data-encoding}{clinical-data-encoding}}. The data cannot be made available due to privacy and ownership reasons.

\footnotesize
\bibliographystyle{unsrt}
\bibliography{bibliography_CIBB_file.bib} 
\normalsize

\end{document}